\documentclass[runningheads]{llncs}
\usepackage{graphicx}
\usepackage{comment}
\usepackage{amsmath,amssymb} 
\usepackage{color}

\usepackage{epsfig}
\usepackage{float}
\usepackage[export]{adjustbox}

\usepackage{multirow}
\usepackage{wrapfig}

\usepackage[pagebackref=true,breaklinks=true,letterpaper=true,colorlinks,bookmarks=false]{hyperref}
\definecolor{darkgreen}{RGB}{64, 191, 128}  

\begin{document}
\pagestyle{headings}
\mainmatter

\title{ProxyNCA++: Revisiting and Revitalizing Proxy Neighborhood Component Analysis} 


\titlerunning{ProxyNCA++: Revisiting and Revitalizing ProxyNCA}
%
\author{Eu Wern Teh\inst{1,2} \and
Terrance DeVries\inst{1,2} \and
Graham W. Taylor\inst{1,2}}
\authorrunning{Teh et al.}
%
\institute{University of Guelph, ON, Canada \and
Vector Institute, ON, Canada \\
\email{\{eteh,terrance,gwtaylor\}@uoguelph.ca}}
\maketitle

\begin{abstract}

We consider the problem of distance metric learning (DML), where the task is to learn an effective similarity measure between images.
We revisit ProxyNCA and incorporate several enhancements. We find that low temperature scaling is a performance-critical component and explain why it works. Besides, we also discover that Global Max Pooling works better in general when compared to Global Average Pooling. Additionally, our proposed fast moving proxies also addresses small gradient issue of proxies, and this component synergizes well with low temperature scaling and Global Max Pooling.
Our enhanced model, called ProxyNCA++, achieves a 22.9 percentage point average improvement of Recall@1 across four different zero-shot retrieval datasets compared to the original ProxyNCA algorithm.
Furthermore, we achieve state-of-the-art results on the CUB200, Cars196, Sop, and InShop datasets, achieving Recall@1 scores of 72.2, 90.1, 81.4, and 90.9, respectively. 

\end{abstract}

\section{Introduction}\label{sec:intro}

Distance Metric Learning (DML) is the task of learning effective similarity measures between examples. It is often applied to images, and has found numerous applications such as visual products retrieval \cite{liu2016deepfashion,song2016deep,Bell2015}, person re-identification \cite{Zheng_2019_CVPR,Wang:2018:LDF:3240508.3240552}, face recognition \cite{Schroff_2015_CVPR}, few-shot learning~\cite{Vinyals:2016:MNO:3157382.3157504,koch2015siamese}, and clustering \cite{7471631}. In this paper, we focus on DML's application on zero-shot image retrieval \cite{movshovitz2017no,wu2017sampling,song2016deep,jacob2019metric}, where the task is to retrieve images from previously unseen classes.

Proxy-Neighborhood Component Analysis (ProxyNCA)~\cite{movshovitz2017no} is a proxy-based DML solution that consists of updatable proxies, which are used to represent class distribution. It allows samples to be compared with these proxies instead of one another to reduce computation. 
After the introduction of ProxyNCA, there are very few works that extend ProxyNCA~\cite{xuan2018deep,sanakoyeu2019divide}, making it less competitive when compared with recent DML solutions~\cite{wang2019multi,jacob2019metric,zhai2019}.

\begin{figure}[H]
  \centering
    \caption{A summary of the average performance on Recall@1 for all datasets. With our proposed enhancements, we improve upon the original ProxyNCA by $22.9$pp, and outperform current state-of-the-art models by $2.0$pp on average.}
    \bigskip
  \includegraphics[width=3.29in, ]{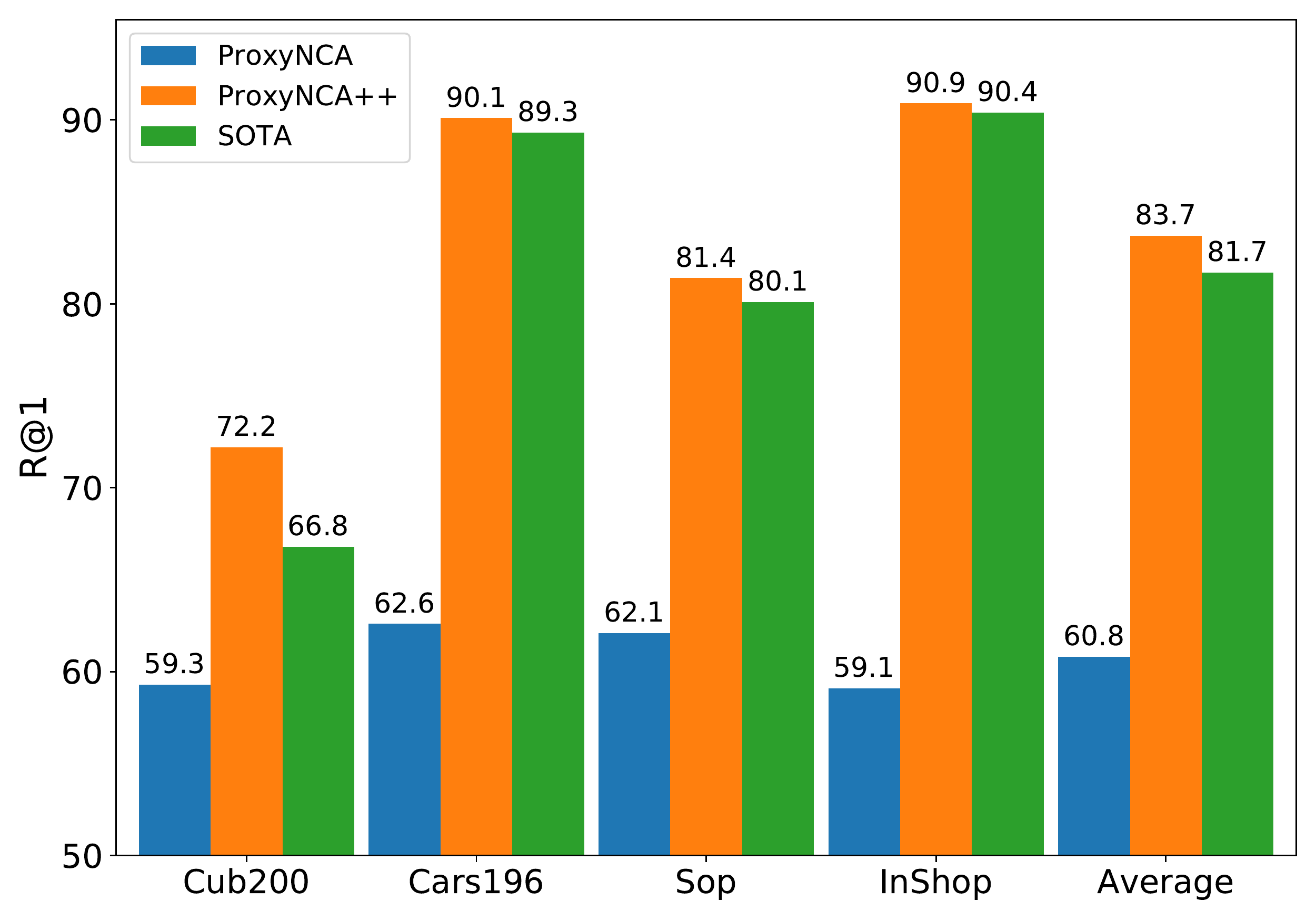}
  \label{fig:summ}
\end{figure}

Our contributions are the following: First, we point out the difference between NCA and ProxyNCA, and propose to use proxy assignment probability which aligns ProxyNCA with NCA~\cite{goldberger2005neighbourhood}. Second, we explain why low temperature scaling works and show that it is a performance-critical component of ProxyNCA. Third, we explore different global pooling strategies and find out that Global Max Pooling (GMP) outperforms the commonly used Global Average Pooling (GAP), both for ProxyNCA and other methods. Fourth, we suggest using faster moving proxies that compliment well with both GMP and low temperature scaling, which also address the small gradient issue due to $L^2$-Normalization of proxies. 
Our enhanced ProxyNCA, which we called ProxyNCA++, has a $22.9$ percentage points of improvement over ProxyNCA on average for Recall@1 across four different zero-shot retrieval benchmarks (performance gains are highlighted in Figure~\ref{fig:summ}). In addition, we also achieve state-of-the-art performance on all four benchmark dataset across all categories.

\section{Related Work}\label{sec:related}
The core idea of Distance Metric Learning (DML) is to learn an embedding space where similar examples are attracted, and dissimilar examples are repelled. 
To restrict the scope, we limit our review to methods that consider image data. There is a large body of work in DML, and it can be traced back to the 90s, where Bromley et al.~\cite{Bromley:1993:SVU:2987189.2987282} designed a Siamese neural network to verify signatures. Later, DML was used in facial recognition, and dimensionality reduction in the form of a contrastive loss \cite{chopra2005learning,hadsell2006dimensionality},
where pairs of similar and dissimilar images are selected, and the distance between similar pairs of images is minimized while the distance between dissimilar images is maximized.

Like contrastive loss, which deals with the actual distance between two images, triplet loss optimizes the relative distance between positive pair (an anchor image and an image similar to anchor image) and negative pair (an anchor image and an image dissimilar to anchor image)~\cite{Chechik:2010:LSO:1756006.1756042}. In addition to contrastive and triplet loss, there is a long line of work which proposes new loss functions, such as angular loss~\cite{Wang_2017_ICCV}, histogram loss~\cite{NIPS2016_6464}, margin-based loss~\cite{wu2017sampling}, and hierarchical triplet loss~\cite{Ge_2018_ECCV}. Wang et al.~\cite{wang2019multi}~categorize this group as paired-based DML.

One weakness of paired-based methods is the sampling process. First, the number of possible pairs grows polynomially with the number of data points, which increases the difficulty of finding an optimal solution. Second, if a pair or triplet of images is sampled randomly, the average distance between two samples is approximately $\sqrt{2}$-away \cite{wu2017sampling}. In other words, a randomly sampled image is highly redundant and provides less information than a carefully chosen one.

In order to overcome the weakness of paired-based methods, several works have been proposed in the last few years. Schroff et al.~\cite{Schroff_2015_CVPR} explore a curriculum learning strategy where examples are selected based on the distances of samples to the anchored images. They use a semi-hard negative mining strategy to select negative samples where the distances between negative pairs are at least greater than the positive pairs. However, such a method usually generates very few semi-hard negative samples, and thus requires very large batches (on the order of thousands of samples) in order to be effective.
Song et al.~\cite{song2016deep} propose to utilize all pair-wise samples in a mini-batch to form triplets, where each positive pair compares its distance with all negative pairs. Wu et al.~\cite{wu2017sampling} proposed a distance-based sampling strategy, where examples are sampled based on inverse $n$-dimensional unit sphere distances from anchored samples. Wang et al.~\cite{wang2019multi} propose a mining and weighting scheme, where informative pairs are sampled by measuring positive relative similarity, and then further weighted using self-similarity and negative relative similarity.

Apart from methods dedicated to addressing the weakness of pair-based DML methods, there is another line of work that tackles DML via class distribution estimation. The motivation for this camp of thought is to compare samples to proxies, and in doing so, reduce computation.  One method that falls under this line of work is the Magnet Loss~\cite{rippel2015metric} in which samples are associated with a cluster centroid, and at each training batch, samples are attracted to cluster centroids of similar classes and repelled by cluster centroids of different classes. Another method in this camp is ProxyNCA~\cite{movshovitz2017no}, where proxies are stored in memory as learnable parameters. During training, each sample is pushed towards its proxy while repelling against all other proxies of different classes. ProxyNCA is discussed in greater detail in Section \ref{sec:proxynca}.

Similar to ProxyNCA, Zhai et al.~\cite{zhai2019} design a proxy-based solution that emphasizes on the Cosine distance rather than the Euclidean squared distance. They also use layer norm in their model to improve robustness against poor weight initialization of new parameters and introduces class balanced sampling during training, which improves their retrieval performance. In our work, we also use these enhancements in our architecture.

Recently, a few works in DML have explored ensemble techniques. 
Opitz et al.~\cite{Opitz_2017_ICCV} train an ensemble DML by reweighting examples using online gradient boosting. 
The downside of this technique is that it is a sequential process. 
Xuan et al.~\cite{xuan2018deep} address this issue by proposing an ensemble technique where ensemble models are trained separately on randomly combined classes.
Sanakoyeu et al.~\cite{sanakoyeu2019divide} propose a unique divide-and-conquer strategy where the data is divided periodically via clustering based on current combined embedding during training. Each cluster is assigned to a consecutive chunk of the embedding, called learners, and they are randomly updated during training. Apart from ensemble techniques, there is recent work that attempts to improve DML in general. Jacob et al.~\cite{jacob2019metric} discover that DML approaches that rely on Global Average Pooling (GAP) potentially suffer from the scattering problem, where features learned with GAP are sensitive to outlier. To tackle this problem, they propose HORDE, which is a high order regularizer for deep embeddings that computes higher-order moments of features.

\section{Methods}\label{sec:method}

\begin{figure}
  \centering
  \caption{
  We show an overview of our architecture, ProxyNCA++, which consists of the original building blocks of ProxyNCA and six enhancements, which are shown in the dashed boxes. ProxyNCA consists of a pre-trained backbone model, a randomly initialized embedding layer, and randomly initialized proxies. The six enhancements in ProxyNCA++ are proxy assignment probability (+prob), low temperature scaling (+scale), class balanced sampling (+cbs), layer norm (+norm), global max pooling (+max) and fast-moving proxies (+fast).
  }
  \bigskip
  \includegraphics[width=3.2in]{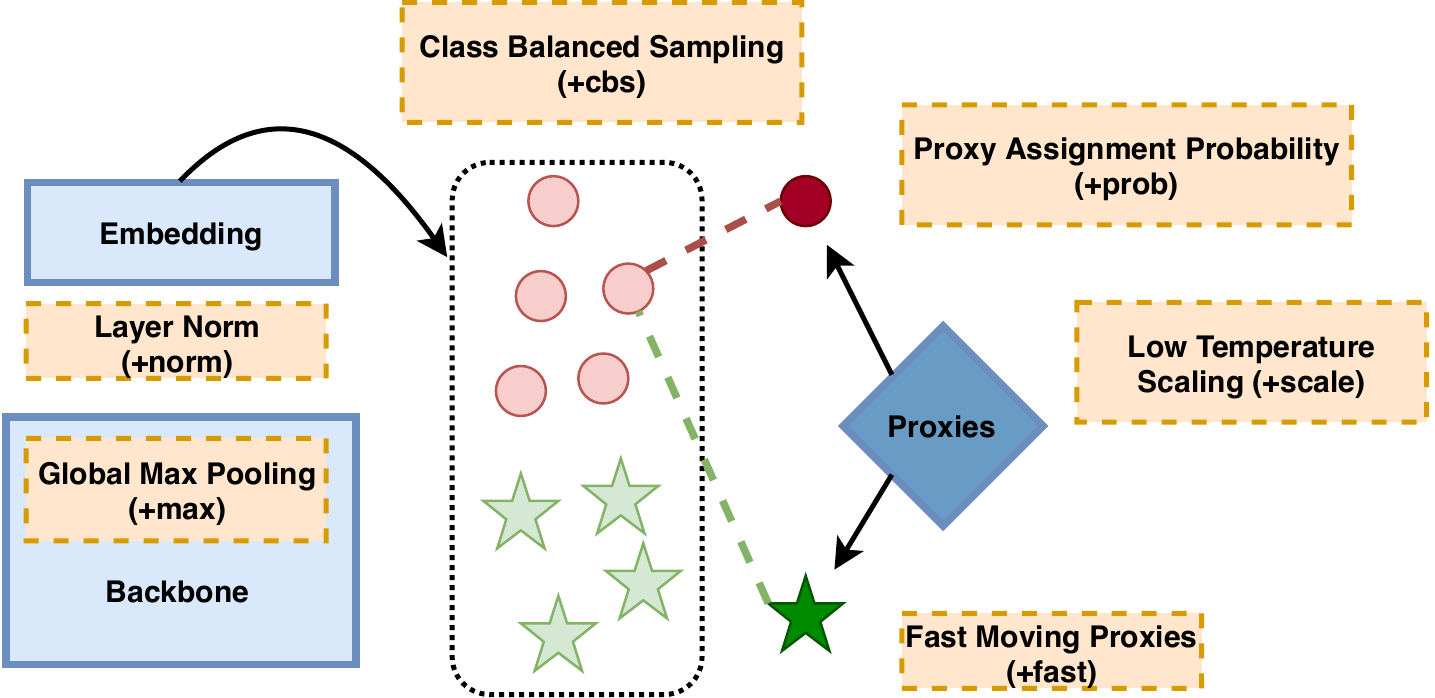}

  \label{fig:pnca_pp}
\end{figure}

In this section, we revisit NCA and ProxyNCA and discuss six enhancements that improve the retrieval performance of ProxyNCA. The enhanced version, which we call ProxyNCA++, is shown in Figure~\ref{fig:pnca_pp}.

\subsection{Neighborhood Component Analysis (NCA)}
Neighborhood Component Analysis (NCA) is a DML algorithm that learns a Mahalanobis distance for k-nearest neighbors (KNN).  Given two points, $x_i$ and $x_j$, Goldberg et al.~\cite{goldberger2005neighbourhood} define $p_{ij}$ as the assignment probability of $x_i$ to $x_j$:
 \begin{equation}\label{eq:4}
 p_{ij} = \frac{-d(x_i, x_j)}{\sum_{k \not\in i} -d(x_i, x_k)}
\end{equation}

\noindent where $d(x_i, x_k)$ is Euclidean squared distance computed on some learned
embedding. In the original work, it was parameterized as a linear mapping, but
nowadays, the method is often used with nonlinear mappings such as feedforward or
convolutional neural networks. Informally, $p_{ij}$ is the probability that
points $i$ and $j$ are said to be ``neighbors''.

The goal of NCA is to maximize the probability that points assigned to the same
class are neighbors, which, by normalization, minimizes the probability that
points in different classes are neighbors:

\begin{equation}\label{eq:nca}
 L_{\text{NCA}} = -\log\left(\frac{\sum_{j \in C_i}^{}\text{exp}(-d(x_i, x_j))}{\sum_{k \not\in C_i}^{}\text{exp}(-d(x_i, x_k))}\right) .\
\end{equation}

Unfortunately, the computation of NCA loss grows polynomially with the number of samples in the dataset. To speed up computation, Goldberg et al.~use random sampling and optimize the NCA loss with respect to the small batches of samples.

\subsection{ProxyNCA}
\label{sec:proxynca}
ProxyNCA is a DML method which performs metric learning in the space of class distributions. It is motivated by NCA, and it attempts to address the computation weakness of NCA by using proxies. 
In ProxyNCA, \emph{proxies} are stored as learnable parameters to faithfully represent classes by prototypes in an embedding space. During training, instead of comparing samples with one another in a given batch, which is quadratic in computation with respect to the batch size, ProxyNCA compares samples against proxies, where the objective aims to attract samples to their proxies and repel them from all other proxies. 

Let $C_i$ denote a set of points that belong to the same class, $f(a)$ be a proxy function that returns a corresponding class proxy, and ${||a||}_2$ be the $L^2$-Norm of vector $a$. For each sample $x_i$, we minimize the distance  $d(x_i, f(x_i))$ between the sample, $x_i$ and its own proxy, $f(x_i)$ and maximize the distance  $d(x_i, f(z))$ of that sample with respect to all other proxies $Z$, where $ f(z) \in Z$ and $z \not \in C_i$.

\begin{equation}\label{eq:1}
 L_{\text{ProxyNCA}} = -\log\left(\frac{\text{exp}\left(-d(\frac{x_i}{||x_i||_2}, \frac{f(x_i)}{||f(x_i)||_2})\right)}{\sum_{f(z)\in Z}^{}\text{exp}\left(-d(\frac{x_i}{||x_i||_2}, \frac{f(z)}{||f(z)||_2})\right)}\right) .\
\end{equation}
\subsection{Aligning with NCA by optimizing proxy assignment probability}

Using the same motivation as NCA (Equation~\ref{eq:4}), we propose to optimize the proxy assignment probability, $P_{i}$. Let  $A$ denote the set of all proxies. For each $x_i$, we aim to maximize $P_{i}$.

\begin{align}\label{eq:pp}
 &P_{i} = \frac{\text{exp}\left(-d(\frac{x_i}{||x_i||_2}, \frac{f(x_i)}{||f(x_i)||_2})\right)}{\sum_{f(a)\in A}^{}\text{exp}\left(-d(\frac{x_i}{||x_i||_2}, \frac{f(a)}{||f(a)||_2})\right)} \\
 &  L_{\text{ProxyNCA++}} = -\log(P_{i})
\end{align}

Since $P_i$ is a probability score that must sum to one, maximizing $P_i$ for a proxy also means there is less chance for $x_i$ to be assigned to other proxies. In addition, maximizing $P_i$ also preserves the original ProxyNCA properties where $x_i$ is attracted toward its own proxy $f(x_i)$ while repelling proxies of other classes, $Z$. It is important to note that in ProxyNCA, we maximize the distant ratio between $-d(x_i, y_j)$ and $\sum_{f(z)\in Z}^{}-d(x_i, f(z))$, while in ProxyNCA++, we maximize the proxy assignment probability, $P_{i}$, a subtle but important distinction. Table~\ref{table:prob_det} shows the effect of proxy assignment probability to ProxyNCA and its enhancements.

\subsection{About Temperature Scaling}

Temperature scaling is introduced in \cite{hinton2015distilling}, where Hinton et al. use a high temperature ($T > 1$) to create a softer probability distribution over classes for knowledge distillation purposes. Given a logit $y_i$ and a temperature variable $T$, a temperature scaling is defined as $q_{i} = \frac{\exp(y_i/T)}{\sum_j \exp(y_j/T)}$. By incorporating temperature scaling to the loss function of ProxyNCA++ in Equation~\ref{eq:pp}, the new loss function has the following form:

\begin{align}\label{eq:ppp}
 &L_{\text{ProxyNCA++}} = -\log\left(\frac{\text{exp}\left(-d(\frac{x_i}{||x_i||_2}, \frac{f(x_i)}{||f(x_i)||_2})*\frac{1}{T}\right)}{\sum_{f(a)\in A}^{}\text{exp}\left(-d(\frac{x_i}{||x_i||_2}, \frac{f(a)}{||f(a)||_2})*\frac{1}{T}\right)} \right)
\end{align}

\begin{figure}
  \centering
  \caption{The effect of temperature scaling on the decision boundary of a Softmax Classifier trained on the two moons synthetic dataset}
  \includegraphics[width=4.5in]{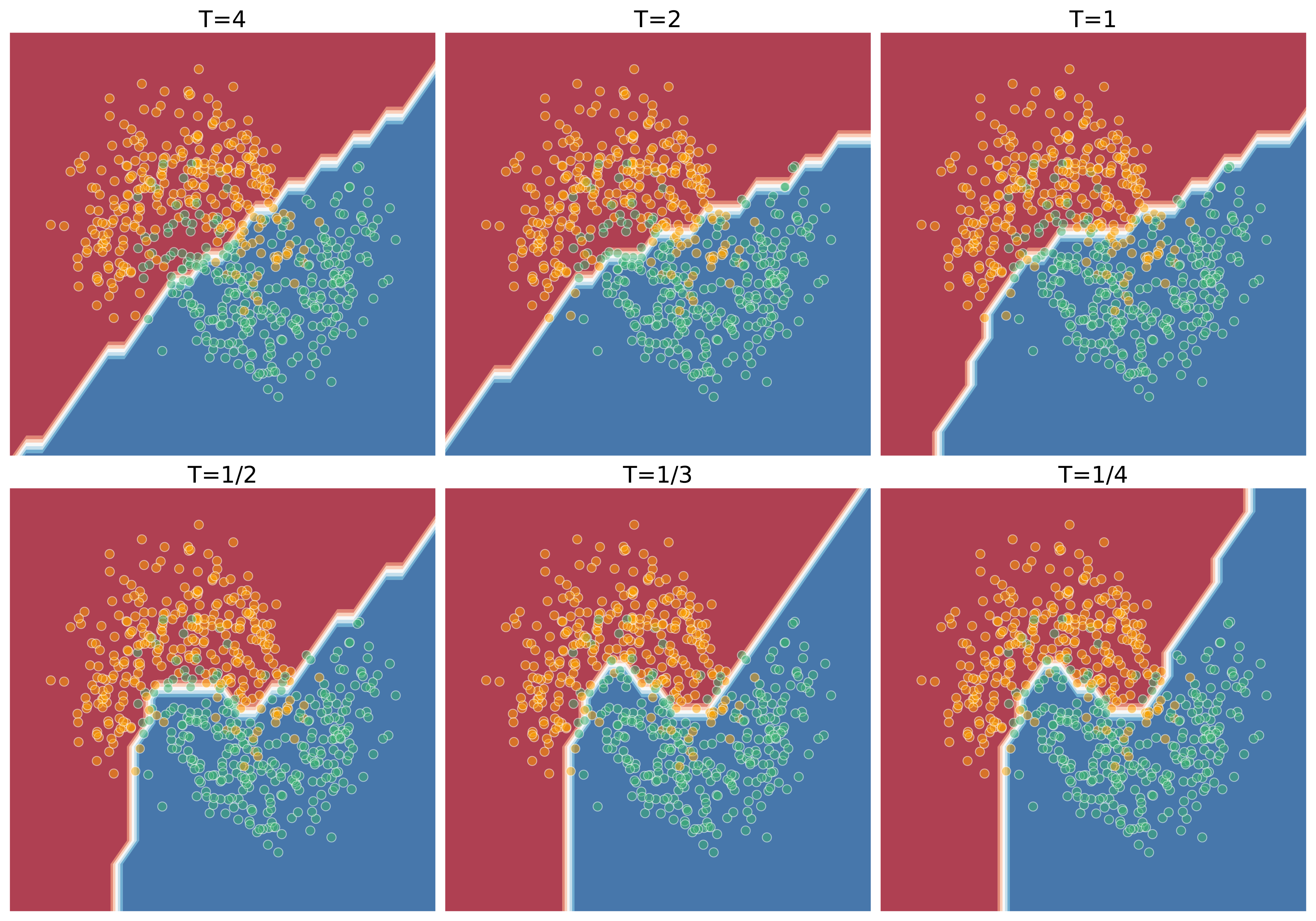}
  \label{fig:temp_scale}
\end{figure}

When $T = 1$, we have a regular Softmax function. As $T$ gets larger, the output of the softmax function will approach a uniform distribution. On the other hand, as $T$ gets smaller, it leads to a peakier probability distribution. Low temperature scaling ($T < 1$) is used in \cite{wu2018improving} and \cite{zhai2019}. In this work, we attempt to explain why low-temperature scaling works by visualizing its effect on synthetic data. In Figure~\ref{fig:temp_scale}, as $T$ gets smaller, the decision boundary is getting more refined and can classify the samples better. In other words, as $T$ becomes smaller, the model can overfit to the problem better and hence generating better decision boundaries.

\begin{figure}
  \centering
  \caption{We show three plots of R@1 with different (a) temperature scales , (b) $k$ values for $K$-Max Pooling and (c) proxy learning rates on on CUB200~\cite{wah2011caltech}. The shaded areas represent one standard deviation of uncertainty. }
  \includegraphics[width=4.8in]{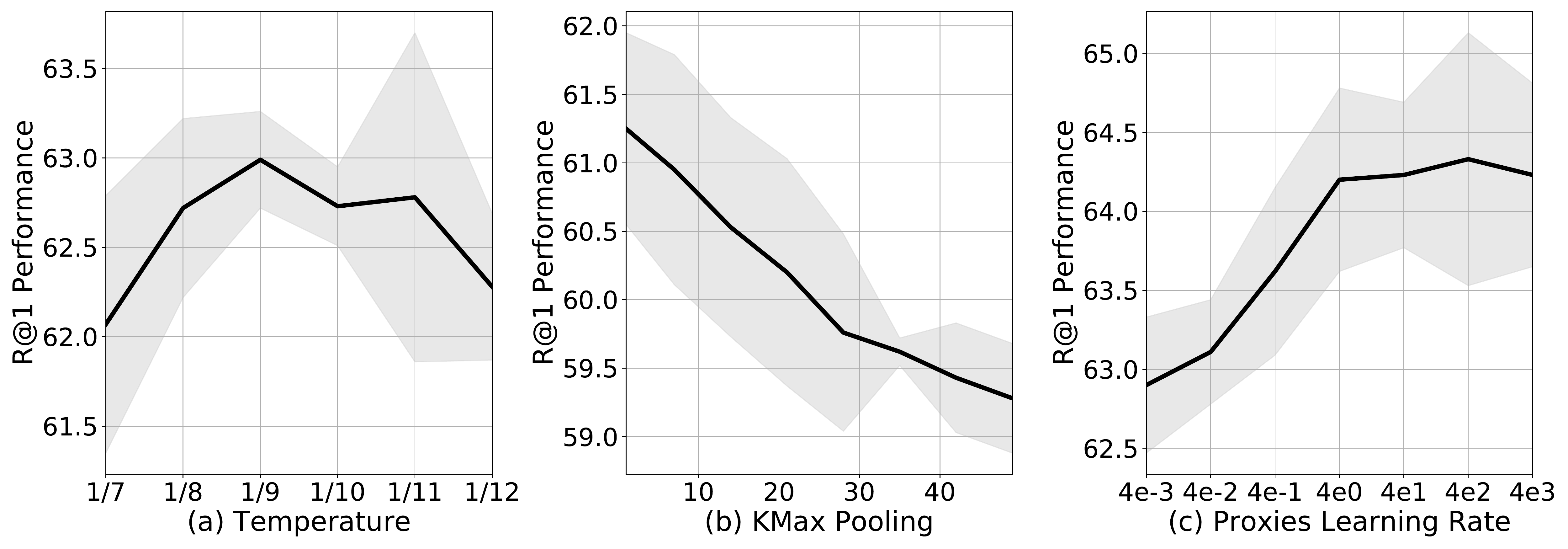}
  \label{fig:ts_kmax}
\end{figure}

In Figure~\ref{fig:ts_kmax} (a), we show a plot of R@1 score with respect to temperature scale on the CUB200 dataset. The highest test average R@1 happens at $T=\frac{1}{9}$. Lowering $T$ beyond this point will allow the model to overfit more to the training set and to make it less generalizable. Hence, we see a drop in test performance.
Table~\ref{table:tempscale_det} shows the effect of low temperature scaling to ProxyNCA and its enhancements.

\subsection{About Global Pooling}
In DML, the de facto global pooling operation used by the community is Global Average Pooling (GAP).  In this paper, we investigate the effect of global pooling of spatial features on zero-shot image retrieval. We propose the use of Global $K$-Max Pooling~\cite{durand2016weldon} to interpolate between GAP and Global Max Pooling (GMP). Given a convolution feature map of $M \times M$ dimension with $E$ channels, $g \in \mathbb{R}^{M \times M \times E}$ and a binary variable, $h_i \in \{0,1\}$, Global K-Max Pooling is defined as:

\begin{equation}\label{eq:kmax}
 \text{Global }k\text{-Max}(g_{\epsilon}) = \max\limits_{h} \frac{1}{k} \sum_{i=1}^{M^2} h_i \cdot g_{\epsilon} \text{\ , s.t.} \sum_{i=1}^{M^2} h_i = k, \forall \epsilon \in E
\end{equation}

When $k=1$, we have GMP, and when $k=M^2$, we have GAP. Figure~\ref{fig:ts_kmax} (b) is a plot of Recall@1 with different $k$ value of Global K-Max Pooling on the CUB200 dataset. There is a negative correlation of 0.98 between $k$ and Recall@1 performance, which shows that a lower $k$ value results in better retrieval performance.

\subsection{About Fast moving proxies}

In ProxyNCA, the proxies, the embedding layer, and the backbone model all share the same learning rate. We hypothesize that the proxies should be moving faster than the embedding space in order to represent the class distribution better. However, in our experiments, we discovered that the gradient of proxies is smaller than the gradient of the embedding layer and backbone model by three orders of magnitude, and this is caused by the  $L^2$-Normalization of proxies. 
To mitigate this problem, we use a higher learning rate for the proxies. 

From our ablation studies in Table~\ref{table:tempscale_det}, we observe that fast moving proxies synergize better with low temperature scaling and Global Max Pooling. We can see a 1.4pp boost in R@1 if we combine fast proxies and low temperature scaling. There is also a 2.1pp boost in the retrieval performance if we combine fast proxies, low temperature scaling, and Global Max Pooling. Figure~\ref{fig:ts_kmax} (c) is a plot of Recall@1 with different proxy learning rates on CUB200.

\subsection{Layer Norm (Norm) and Class Balanced Sampling (CBS)}
The use of layer normalization \cite{vaswani2017attention} without affine parameters is explored by Zhai et al.~\cite{zhai2019}. Based on our experiments, we also find that this enhancement helps to boost performance. Besides, we also use a class balanced sampling strategy in our experiments, where we have more than one instance per class in each training batch. To be specific, for every batch of size $N_b$, we only sample $N_c$ classes from which we then randomly select $\lfloor N_b/N_c\rfloor$ examples. This sampling strategy commonly appears in pair-based DML approaches~\cite{wang2019multi,wu2017sampling,song2016deep} as a baseline and
Zhai et al. is the first paper that uses it in a proxy-based DML method.

\section{Experiments}\label{sec:experiment}

We train and evaluate our model on four zero-shot image retrieval datasets: the Caltech-UCSD Birds dataset~\cite{wah2011caltech} (CUB200), the Stanford Cars dataset~\cite{KrauseStarkDengFei-Fei_3DRR2013} (Cars196), the Stanford Online Products dataset~\cite{song2016deep} (Sop), and the In Shop Clothing Retrieval dataset~\cite{liu2016deepfashion} (InShop). The composition in terms of number of images and classes of each dataset is summarized in Table~\ref{table:dataset}.

\subsection{Experimental Setup}

For each dataset, we use the first half of the original training set as our training set and the second half of the original training set as our validation set. In all of our experiments, we use a two-stage training process. We first train our models on the training set and then use the validation set to perform hyper-parameter tuning (e.g., selecting the best epoch for early stopping, learning rate, etc.). Next, we train our models with the fine-tuned hyper-parameters on the combined training and validation sets (i.e., the complete original training set).

\begin{table}
\centering
\caption{We show the composition of all four zero-shot image retrieval datasets considered in this work. In addition, we also report the learning rates, the batch size, and cbs (class balanced sampling) instances for each dataset during training. The number of classes for the Sop and InShop datasets is large when compared to CUB200 and Cars196 dataset. However, the number of instances per class is very low for the Sop and InShop datasets. In general, ProxyNCA does not require a large batch size when compared to pairs-based DML methods. To illustrate this, we also show the batch sizes used in \cite{wang2019multi}, which is current state-of-the-art among pairs-based methods. Their technique requires a batch size, which is several times larger compared to ProxyNCA++.}
\setlength{\tabcolsep}{3pt}
\begin{tabular}{|lrrccc|ccc|}
\hline
 &  &  &  & batch size & batch size & & & \\
 & images & classes & avg & (ours) & (MS~\cite{wang2019multi}) & Base lr & Proxy lr & cbs\\ \hline
\small{CUB200}& 11,788 & 200 & 58 & 32 & 80      & 4e-3 & 4e2 &4 \\
\small{Cars196}& 16,185 & 196 & 82 & 32 & -      & 4e-3 & 4e2 &4\\
\small{Sop}& 120,053 & 22,634 & 5 & 192 & 1000   & 2.4e-2 & 2.4e2 &3\\
\small{InShop}& 52,712 & 11,967 & 4 & 192 & -    & 2.4e-2 & 2.4e3 &3\\
\hline
\end{tabular}
\label{table:dataset}
\end{table}

We use the same learning rate for both stages of training.
We also set the number of proxies to be the same as the number of classes in the training set. For our experiments with fast proxies, we use a different learning rate for proxies (see Table~\ref{table:dataset} for details). We also use a temperature value of  $\frac{1}{9}$ across all datasets.

In the first stage of training, we use the ``reduce on loss plateau decay'' annealing~\cite{Goodfellow-et-al-2016} to control the learning rate of our model based on the recall performance (R@1) on the validation set. We set the patience value to four epochs in our experiments. We record the epochs where the learning rate is reduced and also save the best epochs for early stopping on the second stage of training.

In all of our experiments, we leverage the commonly used ImageNet~\cite{ILSVRC15} pre-trained Resnet50~\cite{He_2016_CVPR} model as our backbone (see Table~\ref{table:arch} for commonly used backbone architectures). Features are extracted after the final convolutional block of the model and are reduced to a spatial dimension of $1\times 1$ using a global pooling operation. This procedure results in a 2048 dimensional vector, which is fed into a final embedding layer. In addition, we also experiment with various embedding sizes. We observe a gain in performance as we increase the size of the embedding. It is important to note that not all DML techniques yield better performance as embedding size increases. For some techniques such as \cite{wang2019multi,song2016deep}, a larger embedding size hurts performance.

\begin{table}
\centering
\caption{Commonly used backbone architectures for zero-shot image retrieval, with associated ImageNet Top-1 Error \% for each architecture}
\setlength{\tabcolsep}{3pt}
\begin{tabular}{|lcc|}
\hline
Architecture & Abbreviation & Top-1 Error~(\%) \\ \hline
 \small{Resnet18~\cite{He_2016_CVPR}}& R18 &30.24 \\
\small{GoogleNet~\cite{I1}} &I1 &  30.22  \\
\small{Resnet50~\cite{He_2016_CVPR}}&R50&  23.85 \\
\small{InceptionV3~\cite{I3}}& I3 & 22.55 \\
\hline
\end{tabular}
\label{table:arch}
\end{table}

During training, we scale the original images to a random aspect ratio (0.75 to 1.33) before applying a crop of random size (0.08 to 1.0 of the scaled image). After cropping, we resize the images to 256$\times$256. We also perform random horizontal flipping for additional augmentation. During testing, we resize the images to 288$\times$288 and perform a center crop of size 256$\times$256.

\subsection{Evaluation}

 We evaluate retrieval performance based on two evaluation metrics: (a) Recall@K (R@K) and (b) Normalized Mutual Information, $\text{NMI}(\Omega, \mathbb{C}) = \frac{2*I(\Omega, \mathbb{C})}{H(\Omega) + H(\mathbb{C})}$, where $\Omega$ represents ground truth label, $\mathbb{C}$ represents the set of clusters computed by K-means, $I$ stands for mutual information and $H$ stands for entropy. The purpose of NMI is to measure the purity of the cluster on unseen data.

Using the same evaluation protocols detailed in \cite{movshovitz2017no,wang2019multi,jacob2019metric,liu2016deepfashion}, we evaluate our model using unseen classes on four datasets.
The InShop dataset~\cite{liu2016deepfashion} is slightly different than all three other datasets. There are three groupings of data: training set, query set, and gallery set. The query and gallery set have the same classes, and these classes do not overlap with the training set. Evaluation is done based on retrieval performance on the gallery set.

Tables~\ref{table:cub},~\ref{table:cars}, ~\ref{table:sop}, and ~\ref{table:inshop} show the results of our experiments~\footnote[1]{For additional experiments on different crop sizes, please refer to the corresponding supplementary materials in the appendix}.
For each dataset, we report the results of our method, averaged over five runs. We also report the standard deviation of our results to account for uncertainty.
Additionally, we also show the results of ProxyNCA++ trained with smaller embedding sizes (512, 1024).
Our ProxyNCA++ model outperforms ProxyNCA and all other state-of-the-art methods in all categories across all four datasets. Note, our model trained with a 512-dimensional embedding also outperform all other methods in the same embedding space except for The InShop dataset~\cite{liu2016deepfashion}, where we tie in the R@1 category.

\begin{table}[htb]
\centering
\caption{Recall@k for k = 1,2,4,8 and NMI on CUB200-2011~\cite{wah2011caltech}}
\setlength{\tabcolsep}{2pt}
\begin{tabular}{|*8c|}
\hline
R@k & 1 & 2 & 4 & 8 & NMI & Arch & Emb\\ \hline
\small{ProxyNCA\cite{movshovitz2017no}} & 49.2 & 61.9 & 67.9 & 72.4 & 59.5& \small{I1} & \small{128}\\
\small{Margin\cite{wu2017sampling}} & 63.6 & 74.4 & 83.1 & 90.0 & 69.0 & \small{R50}& \small{128}\\
\small{MS~\cite{wang2019multi}}& 65.7 & 77.0 & 86.3 & 91.2 & - & \small{I3}& \small{512}\\
\small{HORDE~\cite{jacob2019metric}}& 66.8 & 77.4 & 85.1 & 91.0 & - & \small{I3}& \small{512}\\
\small{NormSoftMax~\cite{zhai2019}}& 61.3 & 73.9 & 83.5 & 90.0 & - & \small{R50}& \small{512}\\
\small{NormSoftMax~\cite{zhai2019}}& 65.3 & 76.7 & 85.4 & 91.8 & - & \small{R50}& \small{2048}\\
\hline
\small{ProxyNCA} & 59.3$\pm$0.4 & 71.2$\pm$0.3 & 80.7$\pm$0.2 & 88.1$\pm$0.3 & 63.3$\pm$0.5 & \small{R50}& \small{2048}\\
\small{ProxyNCA++} & 69.0$\pm$0.8 & 79.8$\pm$0.7 & 87.3$\pm$0.7 & 92.7$\pm$0.4 & 73.9$\pm$0.5 & \small{R50}& \small{512}\\
\small{ProxyNCA++} & 70.2$\pm$1.6 & 80.7$\pm$1.4 & 88.0$\pm$0.9 & 93.0$\pm$0.4 & 74.2$\pm$1.0 & \small{R50}& \small{1024}\\
\small{ProxyNCA++} & 69.1$\pm$0.5 & 79.6$\pm$0.4 & 87.3$\pm$0.3 & 92.7$\pm$0.2 & 73.3$\pm$0.7 & \small{R50}& \small{2048}\\
\small{(-max, -fast)}& & & & & & &\\
\small{ProxyNCA++} & \textbf{72.2$\pm$0.8} & \textbf{82.0$\pm$0.6} & \textbf{89.2$\pm$0.6} & \textbf{93.5$\pm$0.4} & \textbf{75.8$\pm$0.8} & \small{R50}& \small{2048}\\
\hline
\end{tabular}
\label{table:cub}
\end{table}

\begin{table}[htb]
\centering
\caption{Recall@k for k = 1,2,4,8 and NMI on CARS196~\cite{KrauseStarkDengFei-Fei_3DRR2013}}
\setlength{\tabcolsep}{2pt}
\begin{tabular}{|*8c|}
\hline
R@k & 1 & 2 & 4 & 8 & NMI & Arch & Emb\\ \hline
\small{ProxyNCA~\cite{movshovitz2017no}} & 73.2 & 82.4 & 86.4 & 88.7 & 64.9 &\small{I1}&\small{128}\\
\small{Margin~\cite{wu2017sampling}} & 79.6 & 86.5 & 91.9 & 95.1 & 69.1 & \small{R50}&\small{128}\\
\small{MS~\cite{wang2019multi}}& 84.1 & 90.4 & 94.0 & 96.1 & - &\small{I3}&\small{512}\\
\small{HORDE~\cite{jacob2019metric}}& 86.2 & 91.9 & 95.1 & 97.2 & - & \small{I3}&\small{512}\\
\small{NormSoftMax~\cite{zhai2019}}& 84.2 & 90.4 & 94.4 & 96.9 & - & \small{R50}&\small{512}\\
\small{NormSoftMax~\cite{zhai2019}}& 89.3 & 94.1 & 96.4 & 98.0 & - & \small{R50}&\small{2048}\\
\hline
\small{ProxyNCA}& 62.6$\pm$9.1 & 73.6$\pm$8.6 & 82.2$\pm$6.9 & 88.9$\pm$4.8 & 53.8$\pm$7.0 & \small{R50}&\small{2048}\\
\small{ProxyNCA++}& 86.5$\pm$0.4 & 92.5$\pm$0.3 & 95.7$\pm$0.2 & 97.7$\pm$0.1 & 73.8$\pm$1.0 & \small{R50}&\small{512}\\
\small{ProxyNCA++}& 87.6$\pm$0.3 & 93.1$\pm$0.1 & 96.1$\pm$0.2 & 97.9$\pm$0.1 & 75.7$\pm$0.3 & \small{R50}&\small{1024}\\
\small{ProxyNCA++}& 87.9$\pm$0.2 & 93.2$\pm$0.2 & 96.1$\pm$0.2 & 97.9$\pm$0.1 & 76.0$\pm$0.5 & \small{R50}&\small{2048}\\
\small{(-max, -fast)}& & & & & & & \\
\small{ProxyNCA++}& \textbf{90.1$\pm$0.2} & \textbf{94.5$\pm$0.2} & \textbf{97.0$\pm$0.2} & \textbf{98.4$\pm$0.1} & \textbf{76.6$\pm$0.7} & \small{R50}&\small{2048}\\
\hline
\end{tabular}
\label{table:cars}
\end{table}

\begin{table}[htb]
\centering
\caption{Recall@k for k = 1,10,100,1000 and NMI on Stanford Online Products~\cite{song2016deep}.}
\setlength{\tabcolsep}{2pt}
\begin{tabular}{|*7c|}
\hline
R@k & 1 & 10 & 100 & 1000 & Arch & Emb \\ \hline
\small{ProxyNCA~\cite{movshovitz2017no}} & 73.7 & - & - & - & \small{I1}& \small{128}\\
\small{Margin~\cite{wu2017sampling}} & 72.7 & 86.2 & 93.8 & 98.0 &\small{R50}& \small{128}\\
\small{MS~\cite{wang2019multi}}& 78.2 & 90.5 & 96.0 & 98.7 & \small{I3}& \small{512}\\
\small{HORDE~\cite{jacob2019metric}}& 80.1 & 91.3 & 96.2 & 98.7 & \small{I3}& \small{512}\\
\small{NormSoftMax~\cite{zhai2019}}& 78.2 & 90.6 & 96.2 & - &  \small{R50}& \small{512}\\
\small{NormSoftMax~\cite{zhai2019}}& 79.5 & 91.5 & 96.7 & - &  \small{R50}& \small{2048}\\
\hline
\small{ProxyNCA}& 62.1$\pm$0.4 & 76.2$\pm$0.4 & 86.4$\pm$0.2 & 93.6$\pm$0.3 &  \small{R50}& \small{2048}\\
\small{ProxyNCA++}& 80.7$\pm$0.5 & 92.0$\pm$0.3 & 96.7$\pm$0.1 & 98.9$\pm$0.0 &  \small{R50}& \small{512}\\
\small{ProxyNCA++}& 80.7$\pm$0.4 & 92.0$\pm$0.2 & 96.7$\pm$0.1 & 98.9$\pm$0.0 &  \small{R50}& \small{1024}\\
\small{ProxyNCA++(-max, -fast)}& 72.1$\pm$0.2 & 85.4$\pm$0.1 & 93.0$\pm$0.1 & 96.7$\pm$0.2 &  \small{R50}& \small{2048}\\
\small{ProxyNCA++}& \textbf{81.4$\pm$0.1} & \textbf{92.4$\pm$0.1} & \textbf{96.9$\pm$0.0} & \textbf{99.0$\pm$0.0} &  \small{R50}& \small{2048}\\
\hline
\end{tabular}
\label{table:sop}
\end{table}

\begin{table}[htb]
\centering
\caption{Recall@k for k = 1,10,20,30,40 on the In-Shop Clothing Retrieval dataset~\cite{song2016deep}}
\setlength{\tabcolsep}{2pt}
\begin{tabular}{|*8c|}
\hline
R@k & 1 & 10 & 20 & 30 & 40 &  Arch & Emb \\ \hline
\small{MS~\cite{wang2019multi}}& 89.7 & 97.9 & 98.5 & 98.8 & 99.1 & \small{I3}& \small{512}\\
\small{HORDE~\cite{jacob2019metric}}& 90.4 & 97.8 & 98.4 & 98.7 & 98.9 &  \small{I3}& \small{512}\\
\small{NormSoftMax~\cite{zhai2019}}& 88.6 & 97.5 & 98.4 & 98.8 & -  & \small{R50}& \small{512}\\
\small{NormSoftMax~\cite{zhai2019}}& 89.4 & 97.8 & 98.7 & 99.0 & -  & \small{R50}& \small{2048}\\
\hline
\small{ProxyNCA}& 59.1$\pm$0.7 & 80.6$\pm$0.6 &  84.7$\pm$0.3 & 86.7$\pm$0.4 & 88.1$\pm$0.5  & \small{R50}& \small{2048}\\
\small{ProxyNCA++}& 90.4$\pm$0.2 & 98.1$\pm$0.1  &  98.8$\pm$0.0 & 99.0$\pm$0.1 & 99.2$\pm$0.0  & \small{R50}& \small{512}\\
\small{ProxyNCA++}& 90.4$\pm$0.4 & 98.1$\pm$0.1 &  98.8$\pm$0.1 & 99.1$\pm$0.1 & 99.2$\pm$0.1  & \small{R50}& \small{1024}\\
\small{ProxyNCA++}& 82.5$\pm$0.3 & 93.5$\pm$0.1 &  95.4$\pm$0.2 & 96.3$\pm$0.0 & 96.8$\pm$0.0  & \small{R50}& \small{2048}\\
\small{(-max, -fast)}&  &  &   &  & &  & \\
\small{ProxyNCA++}& \textbf{90.9$\pm$0.3} & \textbf{98.2$\pm$0.0} &  \textbf{98.9$\pm$0.0} & \textbf{99.1$\pm$0.0}& \textbf{99.4$\pm$0.0} & \small{R50}& \small{2048}\\
\hline
\end{tabular}
\label{table:inshop}
\end{table}

\subsection{Ablation Study}\label{sec:ablation}

In Table~\ref{table:cub_det}, we perform an ablation study on the performance of our proposed methods using the CUB200 dataset. The removal of the low temperature scaling component gives the most significant drop in R@1 performance (-10.8pt). This is followed by Global Max Pooling (-3.2pt), Layer Normalization (-2.6pt), Class Balanced Sampling (-2.6pt), Fast proxies (-1.9pt) and Proxy Assignment Probability (-1.1pt).

We compare the effect of the Global Max Pooling (GMP) and the Global Average Pooling (GAP) on other metric learning methodologies \cite{Schroff_2015_CVPR,wu2017sampling,wang2019multi,jacob2019metric} in Table \ref{table:max_other} on CUB200 dataset. The performance of all other models improves when GAP is replaced with GMP, with the exception of HORDE~\cite{jacob2019metric}. In HORDE, Jacob et al.~\cite{jacob2019metric} include both the pooling features as well as the higher-order moment features in the loss calculation. We speculate that since this method is designed to reduce the effect of outliers, summing max-pooled features canceled out the effect of higher-order moment features, which may have lead to sub-optimal performance.

\begin{table}[htb]
\centering
\caption{An ablation study of ProxyNCA++ and its enhancements on CUB200~\cite{wah2011caltech}.  }
\setlength{\tabcolsep}{3pt}
\begin{tabular}{|l|*5c|}
\hline
R@k & 1 & 2 & 4 & 8 & NMI \\
\hline
\small{ProxyNCA++ (Emb: 2048)} & 72.2$\pm$0.8 & 82.0$\pm$0.6 & 89.2$\pm$0.6 & 93.5$\pm$0.4 & 75.8$\pm$0.8 \\
\hspace{0.3cm}\small{-scale}&61.4$\pm$0.4 & 72.4$\pm$0.5 & 81.5$\pm$0.3 & 88.4$\pm$0.5 & 64.8$\pm$0.4 \\
\hspace{0.3cm}\small{-max}& 69.0$\pm$0.6 & 80.3$\pm$0.5 & 88.1$\pm$0.4 & 93.1$\pm$0.1 & 74.3$\pm$0.4 \\
\hspace{0.3cm}\small{-norm}& 69.6$\pm$0.3 & 80.5$\pm$0.5 & 88.0$\pm$0.2 & 93.0$\pm$0.2 & 75.2$\pm$0.4 \\
\hspace{0.3cm}\small{-cbs}& 69.6$\pm$0.6 & 80.1$\pm$0.3 & 87.7$\pm$0.3 & 92.8$\pm$0.2 & 73.4$\pm$0.3 \\
\hspace{0.3cm}\small{-fast}& 70.3$\pm$0.9 & 80.6$\pm$0.4 & 87.7$\pm$0.5 & 92.5$\pm$0.3 & 73.5$\pm$0.9 \\
\hspace{0.3cm}\small{-prob}& 71.1$\pm$0.7 & 81.1$\pm$0.3 & 87.9$\pm$0.3 & 92.6$\pm$0.3 & 73.4$\pm$0.8 \\
\hline
\end{tabular}
\label{table:cub_det}
\end{table}

\section{Conclusion}\label{sec:conclude}
We revisit ProxyNCA and incorporate several enhancements. We find that low temperature scaling is a performance-critical component and explain why it works. Besides, we also discover that Global Max Pooling works better in general when compared to Global Average Pooling. Additionally, our proposed fast moving proxies also addresses small gradient issue of proxies, and this component synergizes well with low temperature scaling and Global Average pooling. 
The new and improved ProxyNCA, which we call ProxyNCA++, outperforms the original ProxyNCA by 22.9 percentage points on average across four zero-shot image retrieval datasets for Recall@1. In addition, we also achieve state-of-art results on all four benchmark datasets for all categories.

\begin{table}[ht]
\centering
\caption{An ablation study of the effect of Proxy Assignment Probability (+prob) to ProxyNCA and its enhancements on CUB200~\cite{wah2011caltech}.}
\setlength{\tabcolsep}{3pt}
\begin{tabular}{|l|c|c|}
\hline
R@1 & without prob & with prob  \\ \hline
\small{ProxyNCA (Emb: 2048)}& 59.3 $\pm$ 0.4 & 59.0 $\pm$ 0.4\\
\hspace{0.3cm}\small{+scale}& 62.9 $\pm$ 0.4 & 63.4 $\pm$ 0.6 \\
\hspace{0.3cm}\small{+scale +norm} & 65.3 $\pm$ 0.7 & 65.7 $\pm$ 0.8\\
\hspace{0.3cm}\small{+scale +max} & 65.1 $\pm$ 0.3 & 66.2  $\pm$ 0.3\\
\hspace{0.3cm}\small{+scale +norm +cbs} & 67.2 $\pm$ 0.8 & 69.1 $\pm$ 0.5 \\
\hspace{0.3cm}\small{+scale +norm +cbs +max}& 68.8 $\pm$ 0.7& 70.3 $\pm$ 0.9 \\
\hspace{0.3cm}\small{+scale +norm +cbs +max +fast} & 71.1 $\pm$ 0.7 & 72.2 $\pm$  0.8 \\
\hline
\end{tabular}
\label{table:prob_det}
\end{table}

\begin{table}[ht]
\caption{An ablation study of the effect of low temperature scaling to ProxyNCA and its enhancements on CUB200~\cite{wah2011caltech}. Without low temperature scaling, three out of six enhancements (in red) get detrimental results when they are applied to ProxyNCA.}
\centering
\setlength{\tabcolsep}{3pt}
\begin{tabular}{|l|c|c|}
\hline
R@1 & without scale & with scale  \\ \hline
\small{ProxyNCA (Emb: 2048)}& 59.3 $\pm$ 0.4 & 62.9  $\pm$ 0.4\\
\hspace{0.3cm}\small{+cbs}& \color{red}{54.8 $\pm$ 6.2} & 64.0 $\pm$ 0.4 \\
\hspace{0.3cm}\small{+prob}& \color{red}{59.0 $\pm$ 0.4} & 63.4 $\pm$ 0.6 \\
\hspace{0.3cm}\small{+norm}& 60.2 $\pm$ 0.6 & 65.3 $\pm$ 0.7 \\
\hspace{0.3cm}\small{+max}& 61.3 $\pm$ 0.7 & 65.1  $\pm$ 0.3\\
\hspace{0.3cm}\small{+fast}& \color{red}{56.3 $\pm$ 0.8} & 64.3  $\pm$ 0.8\\
\hspace{0.3cm}\small{+max +fast} & 60.3 $\pm$ 0.5 & 67.2 $\pm$ 0.5 \\
\hspace{0.3cm}\small{+norm +prob +cbs} & 60.4 $\pm$ 0.7 & 69.1  $\pm$ 0.5\\
\hspace{0.3cm}\small{+norm +prob +cbs +max} & 61.2 $\pm$ 0.7 & 70.3  $\pm$ 0.9\\
\hspace{0.3cm}\small{+norm +prob +cbs +max +fast}& 61.4 $\pm$ 0.4 & 72.2 $\pm$ 0.8 \\
\hline
\end{tabular}
\label{table:tempscale_det}
\end{table}

\begin{table}[hb]
\centering
\caption{An ablation study of ProxyNCA the effect of Global Max Pooling to ProxyNCA and its enhancements on CUB200~\cite{wah2011caltech}. We can see a 2.1pp improvement on average after replacing GAP with GMP. }
\setlength{\tabcolsep}{3pt}
\begin{tabular}{|l|c|c|}
\hline
R@1 & Global Average Pooling & Global Max Pooling  \\ \hline
\small{ProxyNCA (Emb: 2048)}& 59.3 $\pm$ 0.4 & 61.3 $\pm$ 0.7\\
\hspace{0.3cm}\small{+cbs}& 54.8 $\pm$ 6.2 & 55.5 $\pm$ 6.2  \\
\hspace{0.3cm}\small{+prob} & 59.0 $\pm$ 0.4 & 61.2 $\pm$ 0.7 \\
\hspace{0.3cm}\small{+norm} & 60.2 $\pm$ 0.6 & 60.9 $\pm$ 0.9 \\
\hspace{0.3cm}\small{+scale}& 62.9 $\pm$ 0.4 & 65.1 $\pm$ 0.3\\
\hspace{0.3cm}\small{+fast} & 56.3 $\pm$ 0.8 & 60.3 $\pm$ 0.5 \\
\hspace{0.3cm}\small{+scale +fast} & 64.3 $\pm$ 0.8 & 67.2  $\pm$ 0.5\\
\hspace{0.3cm}\small{+norm +prob +cbs} & 60.4 $\pm$ 0.7 & 61.2  $\pm$ 0.7\\
\hspace{0.3cm}\small{+norm +prob +cbs +fast} & 56.2 $\pm$ 0.9 & 61.4 $\pm$ 0.4  \\
\hspace{0.3cm}\small{+norm +prob +cbs +scale} & 69.1 $\pm$  0.5 & 70.3 $\pm$ 0.9  \\
\hspace{0.3cm}\small{+norm +prob +cbs +scale +fast} & 69.0 $\pm$  0.6 & 72.2 $\pm$ 0.8 \\
\hline
\end{tabular}
\label{table:cub_det3}
\end{table}

\begin{table}[htb]
\centering
\caption{Comparing the effect of Global Max Pooling and Global Average Pooling on the CUB200 dataset for a variety of methods. 
}
\setlength{\tabcolsep}{3pt}
\begin{tabular}{|lcccc|}
\hline
Method & Pool & R@1 & Arch & Emb \\ \hline
\small{WithoutTraining} & avg & 45.0 & \small{R50}& \small{2048}\\
 & max & \textbf{53.1} & \small{R50}& \small{2048}\\
\hline
\small{Margin~\cite{wu2017sampling}} & avg & 63.3 & \small{R50}& \small{128}\\
 & max & \textbf{64.3} & \small{R50}& \small{128}\\
\hline
\small{Triplet-Semihard sampling~\cite{Schroff_2015_CVPR}} & avg & 60.5 & \small{R50}& \small{128}\\
 & max & \textbf{61.6} & \small{R50}& \small{128}\\
\hline
\small{MS~\cite{wang2019multi}} & avg & 64.9 & \small{R50}& \small{512}\\
 & max & \textbf{68.5} & \small{R50}& \small{512}\\
 \hline
 \small{MS~\cite{wang2019multi}} & avg & 65.1 & \small{I3}& \small{512}\\
 & max & \textbf{66.1} & \small{I3}& \small{512}\\
\hline
\small{Horde (Contrastive Loss)~\cite{jacob2019metric}} & avg & \textbf{65.1} & \small{I3}& \small{512}\\
 & max & 63.1 & \small{I3}& \small{512}\\
\hline
\end{tabular}

\label{table:max_other}
\end{table}

\clearpage
%
%
\bibliographystyle{ieee_fullname}
\bibliography{eccv}
\appendix{\Large{\textbf{Appendix}}}

\section{A comparison with NormSoftMax~\cite{zhai2019}}

In this section, we compare the differences between ProxyNCA++ and NormSoftMax~\cite{zhai2019}. Both ProxyNCA++ and NormSoftMax are proxy-based DML solutions. By borrowing notations from Equation 6 in the main paper, ProxyNCA++ has the following loss function: 

\begin{align}\label{eq:ppp}
 &L_{\text{ProxyNCA++}} = -\log\left(\frac{\text{exp}\left(-d(\frac{x_i}{||x_i||_2}, \frac{f(x_i)}{||f(x_i)||_2})*\frac{1}{T}\right)}{\sum_{f(a)\in A}^{}\text{exp}\left(-d(\frac{x_i}{||x_i||_2}, \frac{f(a)}{||f(a)||_2})*\frac{1}{T}\right)} \right)
\end{align}

\noindent{And NormSoftMax} has the following loss function:

\begin{align}\label{eq:ns}
 &L_{\text{NormSoftMax}} = -\log\left(\frac{\text{exp}\left(\frac{x_i}{||x_i||_2}^\top \frac{f(x_i)}{||f(x_i)||_2}*\frac{1}{T}\right)}{\sum_{f(a)\in A}^{}\text{exp}\left(\frac{x_i}{||x_i||_2}^\top \frac{f(a)}{||f(a)||_2}*\frac{1}{T}\right)} \right)
\end{align}

The main difference between Equation~\ref{eq:ppp} and~\ref{eq:ns} is the distance function. In ProxyNCA++, we use a euclidean squared distance function instead of cosine distance function. 

\begin{figure}
  \centering
  \caption{A sensitivity study of temperature scaling for NormSoftMax~\cite{zhai2019} without layer norm (norm), class balanced sampling (cbs), and fast proxies (fast). We show a plot of R@1 with different temperature scales on CUB200~\cite{wah2011caltech}. The shaded areas represent one standard deviation of uncertainty. }
  \includegraphics[width=2.5in]{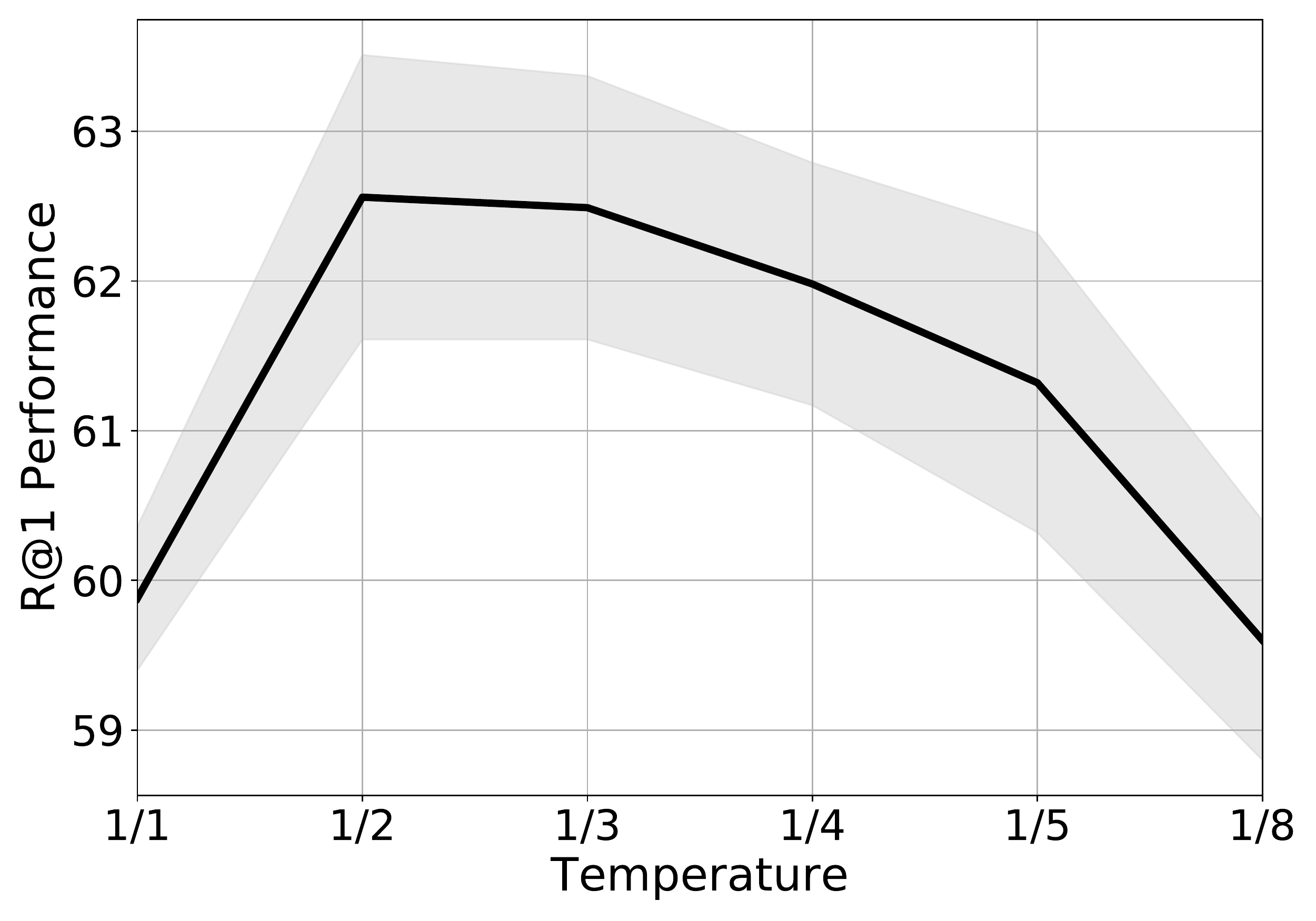}
  \label{fig:n_ts}
\end{figure}

\begin{figure}
  \centering
  \caption{A sensitivity study of proxy learning rate for NormSoftMax~\cite{zhai2019} without layer norm (norm), class balanced sampling (cbs) and with temperature scaling (scale) $T=1/2$. We show a plots of R@1 with different  proxy learning rates on CUB200~\cite{wah2011caltech}. The shaded areas represent one standard deviation of uncertainty. }
  \includegraphics[width=2.5in]{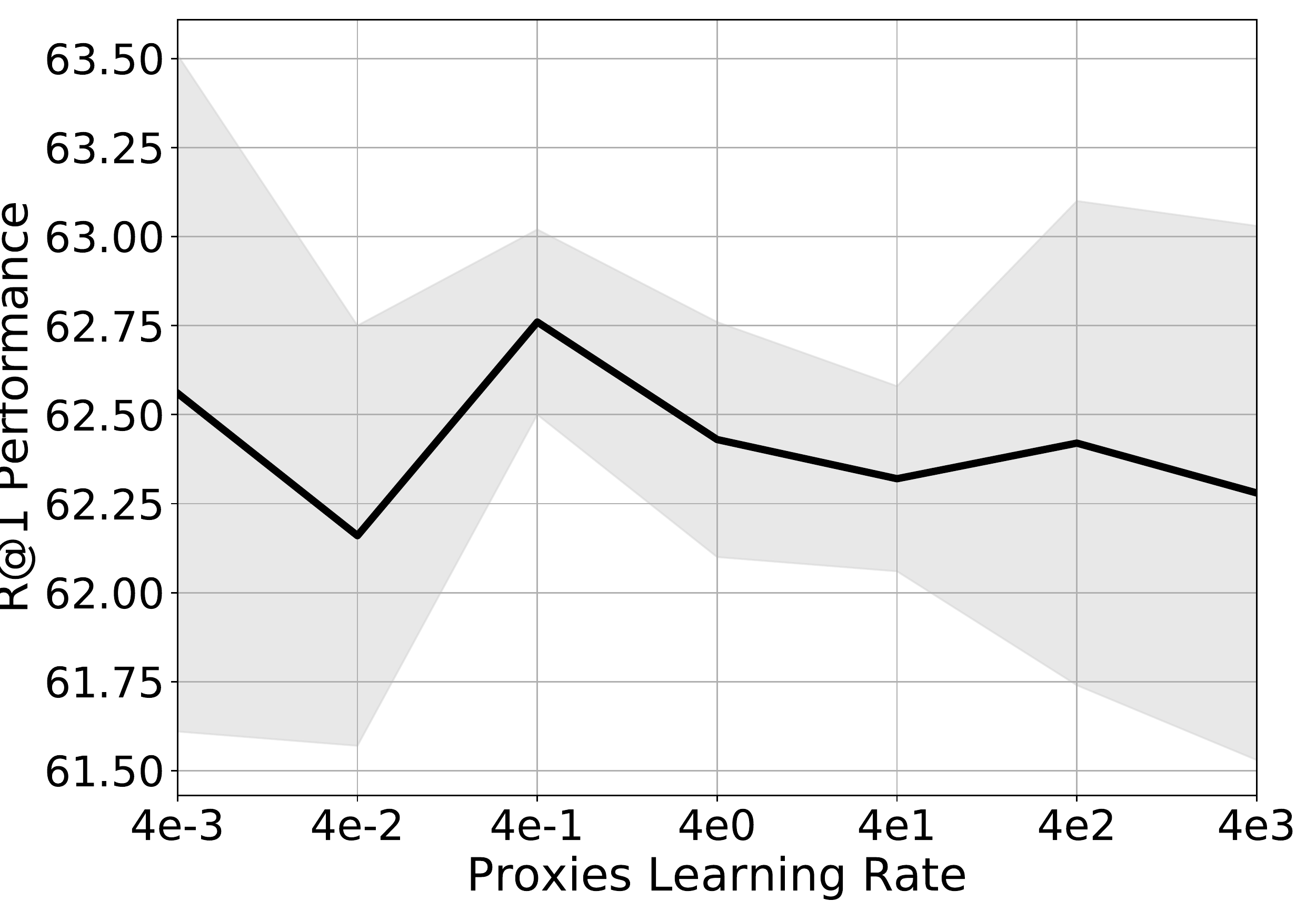}
  \label{fig:n_l}
\end{figure}

Based on our sensitivity studies on temperature scaling and proxy learning rate, we show that NormSoftMax perform best when the temperature scale, $T$ is set to $1/2$ and the proxy learning rate is set to $4e^{-1}$ (see Figure~\ref{fig:n_ts} and~\ref{fig:n_l}).

We perform an ablation study of NormSoftMax in Table~\ref{table:cub_nsm} and \ref{table:cars_nsm}. On the CUB200 dataset, we show that the Global Max Pooling (GMP) component (max) improves NormSoftMax by $2.2$pp on R@1. However, the fast proxies component (fast) reacts negatively with NormSoftMax by decreasing its performance by $0.6$pp. By combining both the GMP and the fast proxies components into NormSoftMax, we see a small increase of R@1 performance (0.6pp).

On the CARS196 dataset, there is a slight increase in R@1 performance (0.3pp) by adding fast proxies component to NormSoftMax. When we add the GMP component to NormSoftMax, we observe an increase of R@1 by 1.1pp. Combining both the GMP and the fast proxies components, there is a 1.0pp increase in R@1 performance.

\begin{table}[htb]
\centering
\caption{A comparison of ProxyNCA++ and NormSoftMax~\cite{zhai2019} on CUB200~\cite{wah2011caltech}.  All models are experimented with embedding size of 2048. For NormSoftMax~\cite{Zheng_2019_CVPR}, we use a temperature scaling of $T=1/2$, a proxy learning rate of $4e^{-1}$ (fast) and learning rates of $4e-3$ for the backbone and embedding layers. It is important to note that, NormSoftMax~\cite{zhai2019}  does not have max pooling and fast proxy component. }
\setlength{\tabcolsep}{3pt}
\begin{tabular}{|l|*5c|}
\hline
R@k & 1 & 2 & 4 & 8 & NMI \\
\hline
\small{ProxyNCA++} & 72.2$\pm$0.8 & 82.0$\pm$0.6 & 89.2$\pm$0.6 & 93.5$\pm$0.4 & 75.8$\pm$0.8 \\
\hspace{0.3cm}\small{-max}& 69.0$\pm$0.6 & 80.3$\pm$0.5 & 88.1$\pm$0.4 & 93.1$\pm$0.1 & 74.3$\pm$0.4 \\
\hspace{0.3cm}\small{-fast}& 70.3$\pm$0.9 & 80.6$\pm$0.4 & 87.7$\pm$0.5 & 92.5$\pm$0.3 & 73.5$\pm$0.9 \\
\hspace{0.3cm}\small{-max -fast}& 69.1$\pm$0.5 & 79.6$\pm$0.4 & 87.3$\pm$0.3 & 92.7$\pm$0.2 & 73.3$\pm$0.7\\
\hline
\small{NormSoftMax} & 65.0$\pm$1.7 & 76.6$\pm$1.1 & 85.5$\pm$0.6 & 91.6$\pm$0.4 & 69.6$\pm$0.8 \\
\small{(+max, +fast)}& & & & &\\
\small{NormSoftMax} & 63.8$\pm$1.3 & 75.9$\pm$1.0 & 84.9$\pm$0.8 & 91.4$\pm$0.6 & 70.8$\pm$1.1 \\
\small{(+fast)}& & & & &\\
\small{NormSoftMax} & 67.6$\pm$0.4 & 78.4$\pm$0.2 & 86.7$\pm$0.4 & 92.2$\pm$0.3 & 71.2$\pm$0.9 \\
\small{(+max)}& & & & &\\
\small{NormSoftMax} & 64.4$\pm$1.1 & 76.1$\pm$0.7 & 85.0$\pm$0.7 & 91.4$\pm$0.3 & 70.0$\pm$1.1 \\
\hline
\end{tabular}
\label{table:cub_nsm}
\end{table}

\begin{table}[H]
\centering
\caption{A comparison of ProxyNCA++ and NormSoftMax~\cite{zhai2019} on CARS196~\cite{KrauseStarkDengFei-Fei_3DRR2013}.  All models are experimented with embedding size of 2048. For NormSoftMax~\cite{Zheng_2019_CVPR}, we use a temperature scaling of $T=1/2$, a proxy learning rate of $4e^{-1}$ (fast) and learning rates of $4e-3$ for the backbone and embedding layers. It is important to note that, NormSoftMax~\cite{zhai2019}  does not have max pooling and fast proxy component. }
\setlength{\tabcolsep}{3pt}
\begin{tabular}{|l|*5c|}
\hline
R@k & 1 & 2 & 4 & 8 & NMI \\
\hline
\small{ProxyNCA++} & 90.1$\pm$0.2 & 94.5$\pm$0.2 & 97.0$\pm$0.2 & 98.4$\pm$0.1 & 76.6$\pm$0.7 \\
\hspace{0.3cm}\small{-max}& 87.8$\pm$0.6 & 93.2$\pm$0.4 & 96.3$\pm$0.2 & 98.0$\pm$0.1 & 76.4$\pm$1.3 \\
\hspace{0.3cm}\small{-fast}& 89.2$\pm$0.4 & 93.9$\pm$0.2 & 96.5$\pm$0.1 & 98.0$\pm$0.1 & 74.8$\pm$0.7\\
\hspace{0.3cm}\small{-max -fast}& 87.9$\pm$0.2 & 93.2$\pm$0.2 & 96.1$\pm$0.2 & 97.9$\pm$0.1 & 76.0$\pm$0.5\\
\hline
\small{NormSoftMax} & 86.0$\pm$0.1 & 92.0$\pm$0.1 & 95.5$\pm$0.1 & 97.6$\pm$0.1 & 68.6$\pm$0.6 \\
\small{(+max, +fast)}& & & & &\\
\small{NormSoftMax} & 85.3$\pm$0.4 & 91.6$\pm$0.3 & 95.5$\pm$0.2 & 97.6$\pm$0.1 & 72.2$\pm$0.7 \\
\small{(+fast)}& & & & &\\
\small{NormSoftMax} & 86.1$\pm$0.4 & 92.1$\pm$0.3 & 95.5$\pm$0.2 & 97.6$\pm$0.2 & 68.0$\pm$0.5  \\
\small{(+max)}& & & & &\\
\small{NormSoftMax} & 85.0$\pm$0.6 & 91.4$\pm$0.5 & 95.3$\pm$0.4 & 97.5$\pm$0.3 & 70.7$\pm$1.1 \\
\hline
\end{tabular}
\label{table:cars_nsm}
\end{table}

\newpage

\section{Two moon classifier}

In Section 3.4 (About Temperature Scaling) in the main paper, we show a visualization of the effect of temperature scaling on the decision boundary of a softmax classifier on a two-moon synthetic dataset. In detail, we trained a two-layers linear model. The first layer has an input size of 2 and an output size of 100. This is followed by a ReLU unit. The second layer has an input size of 100 and an output size of 2. For the synthetic dataset, we use the scikit-learn's~\footnote[2]{\url{https://scikit-learn.org/stable/modules/generated/sklearn.datasets.make_moons.html}} moons data generator to generate 600 samples with noise of 0.3 and a random state of 0. 

\section{Regarding crop size of images}

Image crop size can have a large influence on performance. Current SOTA method~\cite{jacob2019metric} for embedding size 512 uses a crop size of $256\times256$, which we also use for our experiments (See Table~\ref{table:256}). We repeat these experiments with a crop-size of $227\times227$ to make it comparable with older SOTA method~\cite{wang2019multi} (See Table~\ref{table:227}). In this setting, we outperform SOTA for CARS and SOP. We tie on CUB, and we underperform on InShop. However, since no spread information is reported in SOTA~\cite{wang2019multi}, it is hard to make a direct comparison.

\begin{table}[H]
\centering
\caption{A comparison of ProxyNCA++ and the current SOTA~\cite{jacob2019metric} in the embedding size of 512 and a crop size of $256\times256$.}
\setlength{\tabcolsep}{3pt}
\begin{tabular}{|l|*2c|}
\hline
R@k & SOTA~\cite{jacob2019metric} & Ours \\
\hline
\small{CUB} & 66.8 & 69.0$\pm$0.8\\
\small{CARS} & 86.2 & 86.5$\pm$0.4\\
\small{SOP} & 80.1 & 80.7$\pm$0.5\\
\small{InShop} & 90.4 & 90.4$\pm$0.2\\
\hline
\end{tabular}
\label{table:256}
\end{table}

\begin{table}[H]
\centering
\caption{A comparison of ProxyNCA++ and the current SOTA~\cite{wang2019multi} in the embedding size of 512 and a crop size of $227\times227$.}
\setlength{\tabcolsep}{3pt}
\begin{tabular}{|l|*2c|}
\hline
R@k & SOTA~\cite{wang2019multi} & Ours \\
\hline
\small{CUB} & 65.7 & 64.7$\pm$1.6\\
\small{CARS} & 84.2 & 85.1$\pm$0.3\\
\small{SOP} & 78.2 & 79.6$\pm$0.6\\
\small{InShop} & 89.7 & 87.6$\pm$1.0\\
\hline
\end{tabular}
\label{table:227}
\end{table}

\section{Regarding the implementation of baseline}

We follow Algorithm 1 in the original paper~\cite{movshovitz2017no} when implementing our baseline.
In the original paper, there is an $\alpha$ variable that resembles temperature scaling. However, $\alpha$ choice is ambiguous and is used to prove the error bound theoretically. We replicate \cite{movshovitz2017no} on CUB, with embedding size of 64, crop size of 227, and GoogLeNet~\cite{I1} backbone. With the temperature scale, T=1/3, we obtain a R@1 of 49.70, which is close to the reported R@1 49.2. This indicates our implementation of ProxyNCA is correct.

The baseline ProxyNCA is implemented using the same training set up as the proposed ProxyNCA++. As mentioned in our paper (Sec 4.1), we split the original training set into the training (1st half) and validation set (2nd half). We did not perform an extensive sweep of hyperparameters. In our experiment, we first select the best hyperparameter for baseline ProxyNCA (i.e., learning rate [1e-3 to 5e-3]) before adding any enhancements corresponding to ProxyNCA++. We believe that it is possible to obtain better results for both ProxyNCA and ProxyNCA++ with a more extensive sweep of hyperparameters.

\section{Regarding the Global Max Pooling (GMP) vs. Global Average Pooling (GAP)}

\begin{figure}
  \centering
  \caption{Performance summary (R@1) between GMP and GAP on various object sizes (in percent w.r.t. image size) in the CUB dataset.}
  \includegraphics[width=4.3in]{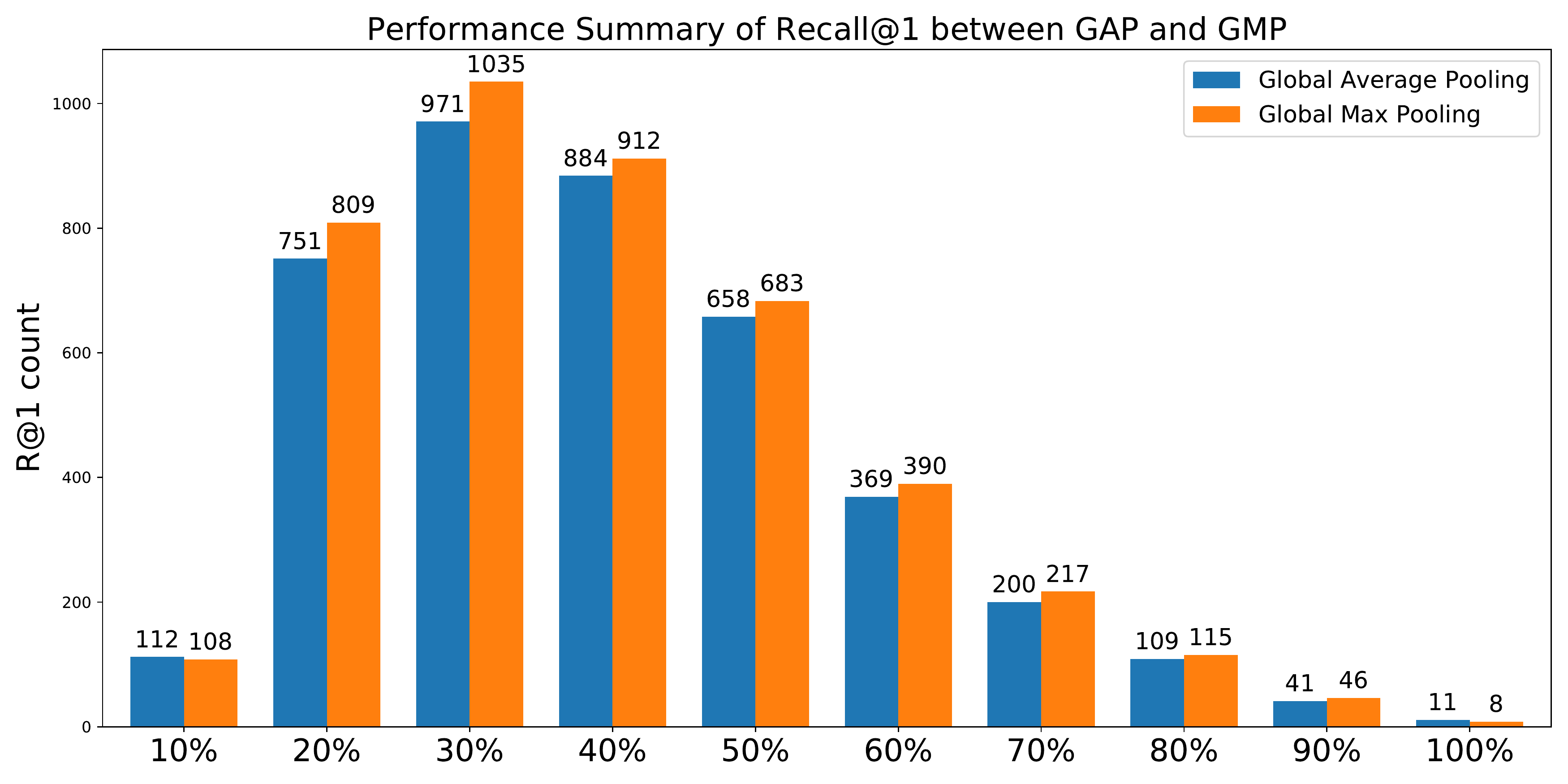}
  \label{fig:gmp_gap_cub}
\end{figure}

In our paper, we show that GMP is better than GAP empirically. However, we could not find any consistent visual evidence as to why GMP works better. We initially hypothesized that GAP was failing for small objects. But after we controlled for object size, we did not observe any consistent visual evidence to support this hypothesis. In Figure~\ref{fig:gmp_gap_cub}, GMP consistently outperform GAP regardless of object size; this evidence disproved our initial hypothesis.  

\newpage
\section{Regarding the computation complexity of ProxyNCA++}

The inference time to compute embeddings for ProxyNCA++ and baseline ProxyNCA will depend on the base architecture. In our experiments, we used a ResNet-50 model as a backbone, so inference time would be comparable to that of a ResNet-50 classifier. There are two differences which have a negligible effect on inference time: (a) The removal of the softmax classification layer, and (b) the addition of layer norm.

As for training time complexity, ProxyNCA++ is comparable to ProxyNCA both theoretically and in terms of runtime. Given a training batch size of B, we only need to compute the distance between each sample w.r.t the proxies, K. After that, we compute a cross-entropy of these distances, where we minimize the probability of a sample being assigned to its own proxy. Therefore the runtime complexity in a given batch is O(BK). 

\end{document}